\pdfoutput=1

\documentclass[11pt]{article}

\usepackage{acl}

\usepackage{times}
\usepackage{latexsym}
\usepackage{tabularx}
\usepackage{multirow}
\usepackage{graphicx}
\usepackage{booktabs}
\usepackage[T1]{fontenc}
\usepackage[ruled]{algorithm2e}
\usepackage{bm}

\usepackage[utf8]{inputenc}

\newcommand{\yblue}[1]{{\color{blue}{{#1}}}}
\newcommand{\yred}[1]{{\color{red}{{#1}}}}
\newcommand{\ygreen}[1]{{\color{Green}{{#1}}}}
\usepackage{siunitx}
\usepackage{makecell}

\usepackage{microtype}
\usepackage{xspace}

\title{FedPETuning: When Federated Learning Meets the Parameter-Efficient Tuning Methods of Pre-trained Language Models}

\author{Zhuo Zhang$^{1,2,}$\footnotemark[1]$^{\;\,}$ \quad Yuanhang Yang$^{1,}$\footnotemark[1] \quad Yong Dai$^{4}$ \quad Qifan Wang$^{5}$ \quad Yue Yu$^{2}$ \\ \textbf{Lizhen Qu}$^{3,}$\footnotemark[2] \quad \textbf{Zenglin Xu}$^{1,2,}$\footnotemark[2] \\
$^{1}$Harbin Institute of Technology, Shenzhen, China\\ 
$^{2}$Peng Cheng Lab, Shenzhen, China \\
$^{3}$Monash University, Melbourne, Australia \\
$^{4}$Tencent, Shenzhen, China \\
$^{5}$Meta AI, CA, USA \\
\{iezhuo17, ysngkil\}@gmail.com \quad daiyongya@outlook.com \quad wqfcr@fb.com \\
yuy@pcl.ac.cn \quad Lizhen.Qu@monash.edu.cn \quad xuzenglin@hit.edu.cn \\
}

\begin{document}
\maketitle
\renewcommand*{\thefootnote}{\fnsymbol{footnote}}
\footnotetext[1]{Equal contribution.}
\footnotetext[2]{Corresponding authors.}
\renewcommand*{\thefootnote}{\arabic{footnote}}

\begin{abstract}

With increasing concerns about data privacy, there is an increasing necessity of fine-tuning pre-trained language models (PLMs) for adapting to downstream tasks located in end-user devices or local clients without transmitting data to the central server. This urgent necessity therefore calls the research of investigating federated learning (FL) for PLMs. However, large PLMs bring the curse of prohibitive communication overhead and local model adaptation costs for the FL system. To this end, we investigate the parameter-efficient tuning (PETuning) of PLMs and develop a corresponding federated benchmark for four representative PETuning methods, dubbed FedPETuning.
Specifically, FedPETuning provides the first holistic empirical study of representative PLMs tuning methods in FL, covering privacy attacks, performance comparisons, and resource-constrained analysis. Intensive experimental results have indicated that FedPETuning can efficiently defend against privacy attacks and maintains acceptable performance with reducing heavy resource consumption. The open-source code and data are available at \url{https://github.com/SMILELab-FL/FedPETuning}.

\end{abstract}

\section{Introduction}
\label{sec:intro}

\textbf{P}re-trained \textbf{L}anguage \textbf{M}odels (PLMs), such as BERT~\cite{devlin2018bert} and RoBERTa~\cite{liu2019roberta}, have demonstrated exceptional performance on a multitude of natural language processing (NLP) benchmarks (e.g., GLUE~\cite{radford2019language}). Consequently, PLMs have become \textit{de facto} backbones for real-world applications. 
Generally, in most real-world NLP applications, PLMs are centrally fine-tuned on the enormous quantity of data collected from individual customers, small businesses, or large enterprises~\cite{qu2021natural}. However, with the rising privacy concerns and the enactment of data protection laws\footnote{Such as the EU’s GDPR
or the US’s HIPAA.}, enterprises or institutions are not allowed to collect data from end devices or local clients to a centralized server for fine-tuning PLMs.



To break this barrier, federated learning~\cite{konevcny2016federated,mcmahan2017communication} (FL) has emerged as a privacy-aware technique designed to collaboratively train models without transmitting the numerous end-user or client data to a centralized place. In FL, the decentralized clients only need to periodically compute and send model information (i.e., parameters or gradients) to a server which is responsible for aggregating them to produce a global model. With the notion of privacy preserving, FL is appealing for privacy-sensitive NLP applications~\cite{sui2020feded,basu2021privacy}, as in the case of healthcare~\cite{ge2020fedner}, finance~\cite{long2020federated}, and mobile keyboard~\cite{ji2019learning}.

\begin{figure}[t]
\centering
\includegraphics[trim={0cm 8cm 0cm 0cm}, clip, scale=0.45]{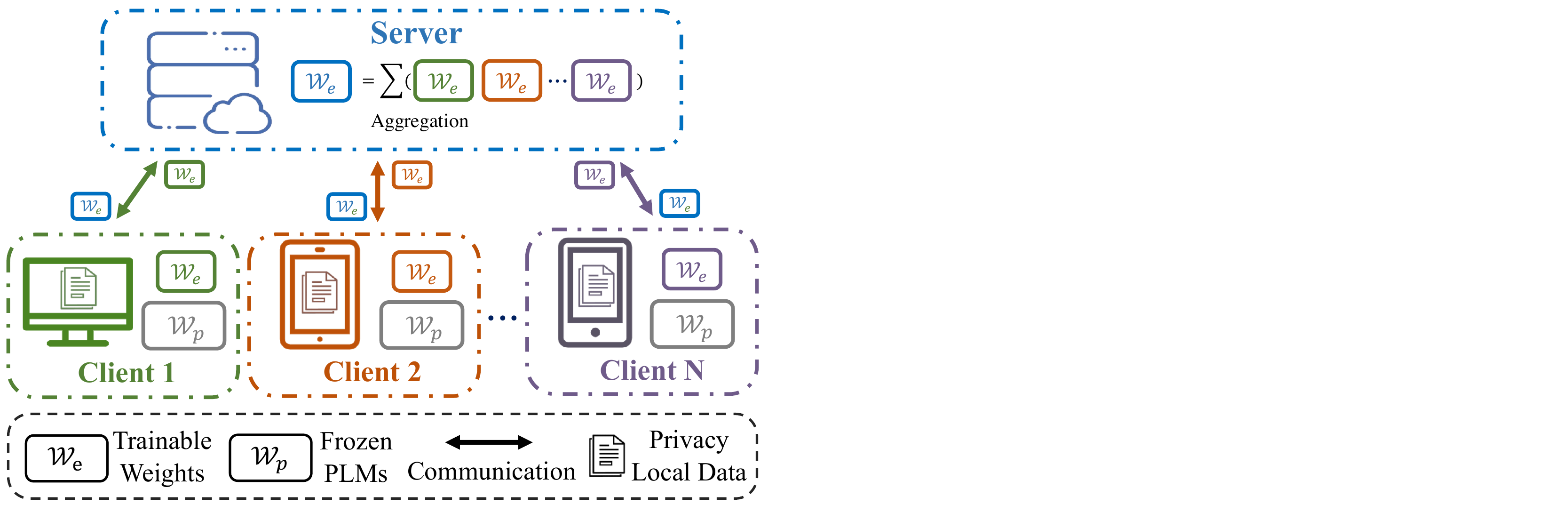}
\caption{An overview of FedPETuning where a client exchanges a light amount of parameters of PLMs with the server while keeping most parameter frozen.}
\label{fig:overview} 
\vspace{-0.4cm}
\end{figure}

Although fine-tuning PLMs in FL, namely FedFT, presents promising opportunities, there are two significant challenges that cannot be overlooked, including (1) communication overhead in the FL system, and (2) computational and storage costs for local clients. Fine-tuning such PLMs usually requires distributed clients and services high-frequently exchange model gradients or parameters which are usually in the scale of millions or even billions. The limited communication bandwidth\footnote{For example, the communication bandwidth between clients and server is constrained from a hundred Kbps to a few Mbps in most situations~\cite{sui2020feded}.} in the FL system may cause excessive delays during frequent uploading and downloading phases of the federated fine-tuning procedure.
Meanwhile, it is often impractical for local clients to fine-tune the entire PLMs because of their limited computing resources. Moreover, fully fine-tuning PLMs is extremely memory-intensive when a local client wants to store different instances for different tasks~\cite{DBLP:conf/iclr/HuSWALWWC22}. 
As such, it is imperative to explore suitable PLMs-empowered FL methods under resource constraints (i.e., communication, parameters adaption, and storage).

To this end, we investigate parameter-efficient tuning (PETuning) methods of PLMs under the FL setting. PETuning methods, such as adapter tuning~\cite{houlsby2019parameter}, prefix tuning~\cite{li-liang-2021-prefix}, LoRA~\cite{DBLP:conf/iclr/HuSWALWWC22}, BitFit~\cite{zaken2021bitfit}, freeze most parameters of PLMs and update only a few additional parameters or a part of the original model parameters for downstream tasks~\cite{ding2022delta}. These properties make PETuning methods potentially appealing to satisfy the resource constraints since local clients just need to tune lightweight parameters and communicate with the server for updates. Nevertheless, it is crucial to address pertinent concerns. First, as the performance of federated models is greatly affected by the data heterogeneity~\cite{kairouz2021advances} among clients, it is not yet known that PETuning methods can achieve acceptable performance in FL with such challenging data heterogeneity. Second, as private data could be recovered from model gradients uploaded by clients via gradient inversion attacks~\cite{zhu2019deep}, it is unclear whether PETuning methods that upload only partial parameters of the entire model can resist gradient inversion attacks. 

To address these concerns, we present the framework of Federated Parameter-Efficient Tuning (named FedPETuning for short), as illustrated in Figure~\ref{fig:overview}, and conduct in-depth experiments of various PETuning methods under the FL setting, measuring privacy-preserving capability, performance, and resource costs. Intensive experimental results on the GLUE benchmark reveal that (1) FedPETuning can reduce the considerable resource costs in FL settings  while still achieving acceptable performance (e.g., FedAP reduces 97.4\% of communication with only 0.7\% performance degradation compared with FedFT) as shown in Table \ref{tab:main_results}, and (2) FedPETuning has an appealing ability to defend against gradient inversion attacks, which can reduce the prevision of recovered words by an average of 40.7\% compared to FedFT, as shown in Section \ref{exp:privacy}. In summary, the major contributions of this paper are shown as follows:
\begin{itemize}
    \item FedPETuning is the first benchmark to provide a holistic review of PETuning methods for PLMs under FL settings, covering privacy attacks, performance comparisons, and resource-constrained analysis. 

    \item FedPEtuning can serve as a suite of baselines for efficient-parameter tuning of PLMs in a federated setting, and guide the community to design FL-tailored efficient parameter tuning algorithms. 
    
    
\end{itemize}

Our research findings demonstrate the potential of combining large PLMs with FL, providing a promising training paradigm for privacy-preserving learning in the era of large language models \cite{ouyang2022training,openai2023gpt4}.

%






\section{Related Work}
\label{sec:related}
\paragraph{Federated Learning} 
Federated learning~\cite{konevcny2016federated,mcmahan2017communication} (FL), a widespread distributed learning technique used in privacy-sensitive tasks, has been hindered by not Independently and Identically Distributed (non-IID)~\cite{kairouz2021advances}, which results in accuracy discrepancies compared to centralized training. Extensive optimization studies have been conducted to address the non-IID issue, including data optimization~\cite{DBLP:journals/corr/abs-1806-00582}, model updating optimization~\cite{DBLP:journals/corr/abs-2010-05958}, and model training optimization~\cite{DBLP:journals/corr/abs-1812-06127}. Recently, some work has tried to solve this problem from a pre-trained model initialization perspective~\cite{chen2022pre,nguyen2022begin} and transformer model structure~\cite{qu2022rethinking}. \citet{weller2022pretrained} experimentally show that using PLMs could reduce non-IID adverse effects and narrow down its accuracy gap to centralized learning. However, the significant communication overhead of large-scale PLMs are less considered in FL systems, leading to slow and impractical federated training in real-world tasks. Additionally, PLMs can pose challenges for local clients with limited hardware capabilities for computation and storage. In contrast, our study investigates PLMs' training in the FL context under resource constraints.

\paragraph{Injecting parameters-efficient tuning methods into federated learning}
Parameter-efficient tuning (PETuning) seeks to keep most parameters of PLMs frozen and fine-tune only additional lightweight parameters or a fraction of the parameters for downstream tasks~\cite{houlsby2019parameter,li-liang-2021-prefix,DBLP:conf/iclr/HuSWALWWC22,zaken2021bitfit}. With this trait, PETuning methods can be utilized to mitigate the communication overhead in FL, which primarily relies on the size of model update parameters. In the field of computer vision, \citet{sun2022exploring} present the FedPEFT framework by injecting three PETuning methods (i.e., Bais, Adapter, Prompt) of the visual pre-trained models into FL, and find that lightweight PETuning methods in FL can significantly reduce the communication burden while maintaining performance and performing better in different FL settings. Meanwhile, \citet{chen2022fedtune} extend PETuning methods to the visual language model in FL and show that PETuning can facilitate a fast convergence rate. However, these studies ignore the important privacy attack issue existing in FL.  With increasing attention to privacy concerns, validation of the privacy-preserving capabilities of federated PETuning methods is paramount and facilitates their practical deployment in real-world scenarios.

In the context of NLP, \citet{zhao2022reduce} first explore the effect of prompt-tuning under the FL setting and achieve acceptable performance results compared with fine-tuning. \citet{xu2022training} show that it is possible to train large vocabulary language models while preserving accuracy and privacy by adopting the low-rank adaptation~\cite{DBLP:conf/iclr/HuSWALWWC22}. However, there has been no comprehensive investigation into the FL performance of the PETuning method for PLMs. Our research aims to bridge this gap and provide access to our code and data to inspire further exploration of the potential of this new paradigm for efficient federated NLP.

\section{Federated Parameter-Efficient Tuning}
\label{sec:method}
In this section, we present how different PETuning methods work, followed by the training process of FedPETuning. 

\subsection{PETuning Methods}
Denote the original PLM parameters by $\mathcal{W}_p=\{w_1, w_2, ..., w_N\}$ and the updated parameters by $\mathcal{W}_p^{'}=\{w_1^{'}, w_2^{'}, ..., w_M^{'}\}$ after training on the dataset $\mathcal{D}$. Define $\mathcal{W}_e$ as trainable model parameters. In vanilla fine-tuning, $|\mathcal{W}_e|$ are equal to the number of original model parameters and $N=M$, where $|\cdot|$ refers to the number of parameters. In the PETuning methods, most parameters of the PLM keep frozen and only a few added parameters or a part of the original model parameters are updated, that is, $M \geq N$ and $|\mathcal{W}_e| \ll N$. Following the taxonomy of \citet{ding2022delta}, PETuning methods could be divided into three groups, i.e., addition-based methods, specification-based methods, and reparameterization-based methods.

\textit{Addition-based methods} introduce new trainable parameters into the frozen PLMs. These methods have two representative branches: \textbf{Adapter} tuning and \textbf{Prompt} tuning. The adapter tuning proposed by \citet{houlsby2019parameter} inserts adapter layers to vanilla Transformers. Specifically, two adapter layers are inserted into each Transformer block, wherein each adapter layer contains a down-projection and an up-projection. Given the input feature ${h} \in \mathcal{R}^{d}$, the down-projection $\mathcal{W}_{d} \in \mathcal{R}^{d \times r}$ projects the input $h_{in}$ to a $r$-dimensional space and the up-projection $\mathcal{W}_{u} \in \mathcal{R}^{r \times d}$ maps back to the input size. Mathematically, the computation process is as follows,
\begin{equation}
    \mathbf{h} \leftarrow \mathcal{W}_u^{T} f\left(\mathcal{W}_d^{T}\mathbf{h}\right),
\end{equation}
\noindent where $f(\cdot)$ is the nonlinear activation function. In this strategy, adapter tuning could only fine-tune adapter layers $\mathcal{W}_e=\{\mathcal{W}_u, \mathcal{W}_d\}$ (about 0.5\% to 8\% parameters of the whole model) during the tuning process while keeping parameters of PLMs frozen. 

On the contrary, prompt-tuning~\cite{li-liang-2021-prefix, DBLP:journals/corr/abs-2110-07602} adds extra parameters without modifying the model architecture. Prompt-tuning converts  the training objective of downstream tasks into a form similar to the pre-trained stage
\cite{devlin-etal-2019-bert, DBLP:journals/corr/abs-1907-11692}, by attaching trainable vectors $\mathcal{P}$, namely prompt, to the original input. During model training, prompt-tuning only adapts light-weight prompt vectors $\mathcal{W}_e=\{\mathcal{P}\}$, which scale to within 10\% of the number of PLMs parameters.

\SetAlFnt{\small}
\SetAlCapFnt{\small}
\SetAlCapNameFnt{\small}
\begin{algorithm}
\DontPrintSemicolon
\SetKwInOut{Parameters}{Parameters}

\noindent \colorbox[gray]{0.95}{
\begin{minipage}{0.85\linewidth}
\textbf{Parameters:} Client set $\mathcal{C}$; Communication round $\mathcal{T}$; Local epoch number $\mathcal{E}$; The PLMs parameters $\mathcal{W}_p$; The local trainable and efficient parameters $\mathcal{W}_e$ and the local dataset $\mathcal{D}_k$ of the $k$-th client; Local PETuning Method $\mathbf{P}$;
\end{minipage}
}

\noindent \colorbox[gray]{0.95}{
\begin{minipage}{0.85\linewidth}
\textbf{Before Training:} Initialize $\mathcal{W}_e^0$ on the server and $\mathcal{W}_p$ on each client in $\mathcal{C}$.
\end{minipage}
}

\noindent \colorbox[rgb]{1, 0.95, 1}{
\begin{minipage}{0.85\linewidth}
\textbf{ServerGlobalAggregation:}

\For{each communication round $t=1$ to $\mathcal{T}$} {
    $\mathcal{C}^{t} \leftarrow$ (randomly sample K clients from $\mathcal{C}$)\;
    \For{each user $k \in \mathcal{C}^{t}$ \textbf{in parallel}}{
        ClientLocalTuning($k$, $\mathcal{W}_e^{t-1}$)\;
    }
    Receive local updated parameters ${\mathcal{W}_{e}^{k,t}}$
    
    Perform global aggregation by Eq.~(\ref{eq:aggregation})\;
}
\end{minipage}
}

\colorbox[rgb]{0.95, 0.98, 1}{
\begin{minipage}{0.85\linewidth}
\SetKwFunction{FUser}{}
    \SetKwProg{Fn}{ClientLocalTuning}{:}{}
    \Fn{\FUser{$k$, $\mathcal{W}_{e}^{t}$}}{
        $\mathcal{W}^{t} \leftarrow$ (assemble $\mathcal{W}_e^t$ and $\mathcal{W}_p$)\;
        \For{epoch $e=1$ to $\mathcal{E}$} {
        ${\mathcal{W}_{e}^{k,t+1}} \leftarrow  \mathbf{P}(\mathcal{D}_k, \mathcal{W}^{t}$)\;
    }
    \textbf{Send} ${\mathcal{W}_{e}^{k,t+1}}$ to the server
}
\end{minipage}
}
\caption{Training process of FedPETuning}
\label{algorithm}
\end{algorithm}  

\textit{Specification-based methods} aim to fine-tune a fraction of the parameters while keeping others frozen. In particular, \citet{zaken2021bitfit} propose \textbf{BitFit} and empirically demonstrate that only tuning the bias terms $\mathcal{W}_e=\{\mathbf{b}_{(\cdot)}^{\ell,(\cdot)}\}$ of PLMs could still achieve competitive performance. 

\textit{Reparameterization-based methods} argue that the PLMs' adaptions can be re-parameterized into optimization within a low-dimensional subspace~\cite{aghajanyan2020intrinsic}. Based on this hypothesis, \textbf{LoRA}~\cite{DBLP:conf/iclr/HuSWALWWC22} optimizes the low-rank decomposition for weight update matrices $\Delta \mathcal{W}$ during model training. For a pretrained weight matrix $\mathcal{W} \in \mathcal{R}^{d \times k}$, we have
\begin{equation}
    \mathcal{W}+\Delta \mathcal{W}=\mathcal{W}+\mathcal{B A},
\end{equation}
\noindent where $\mathcal{B} \in \mathcal{R}^{d \times r}$, $\mathcal{A} \in \mathcal{R}^{r \times k}$, and the rank $r \ll min(d, k)$. In this way, LoRA could make training more efficient with less than 1\% trainable parameters $\mathcal{W}_e=\{\mathcal{B}, \mathcal{A}\}$ and match the fine-tuning performance.

\subsection{FedPETuning}
Our work considers a conventional FL system that comprises a central server responsible for managing participated clients for local training and distributing the shared model parameters. Instead of communicating all cumbersome PLMs parameters, FedPETuning resorts to PETuning methods for exchanging lightweight parameters, as illustrated in Figure~\ref{fig:overview}. 

Before training, FedPETuning initializes the backbone PLM with $\mathcal{W}_p$ and the PETuning method $\mathbf{P}$ with efficient-parameter $\mathcal{W}_e$. Then global aggregation in the server and local updating in local clients are executed alternately. 

\textbf{Server Global Aggregation.} The third-party service first randomly selects $K$ clients from $\mathcal{C}$ and distributes trainable parameter $\mathcal{W}_e$ to these chosen clients. 
Then the server performs the federated aggregation based on the received parameters $\mathcal{W}_e^{k,t}$ from clients, and updates the $\mathcal{W}_e$ by
\begin{equation}
\mathcal{W}_e^{t+1}=\sum_{k=1}^K \frac{\left|\mathcal{D}_k\right|}{\sum_{k=1}^K\left|\mathcal{D}_k\right|} \mathcal{W}_e^{k,t}. \label{eq:aggregation}
\end{equation}

\textbf{Client Local Tuning.} When selected clients download the trainable parameters, they assemble a whole model with lightweight parameters $\mathcal{W}_e$ and local PLM parameters $\mathcal{W}_p$. Then, selected clients train the assembled model with local private data $\mathcal{D}_k$. After local training, the $k$-th client sends its updated efficient parameters $\mathcal{W}^k_e$ to the central server for federated aggregation.

The training process described above is repeated until a specific criterion (e.g., the maximum number of communication rounds $\mathcal{T}$) is satisfied. This process can be summarized in Algorithm~\ref{algorithm}.

\section{Experiments}
\label{sec:exp}
In this section, we conduct extensive experiments for evaluating the performance of PETuning in the FL setting, covering privacy attacks (see Section \ref{exp:privacy}), performance comparisons (see Section \ref{exp:performance}), and resource-constrained analysis (see Section \ref{exp:cost}). Besides, we also provide an in-depth ablation analysis of FedPETuning in terms of data heterogeneity (see Section \ref{exp:niid}), local training epochs (see Section \ref{exp:epoch}), and different FL scenarios (see Section \ref{exp:scenario}). 

\begin{table}[]
\centering
\small
\begin{tabular}{l|c|c|c|c}
\toprule
\textbf{Task} & \textbf{\# Train} & \textbf{\# Dev.} & \textbf{\# Test} & \textbf{Metric}  \\ \midrule
RTE       & 2,241 &  249 &  277  & Accuracy \\
MRPC      & 3,301 &  367 &  408  & F1 Score \\ 
SST-2     & 66,675 &  674 &  872  & Accuracy \\
QNLI      & 103,695 &  1,048 &  5,463 & Accuracy \\
QQP       & 360,210 &  3,639 &  40,430 & Accuracy \\ 
MNLI   & 388,774 &  3,928 &  9,815 & Accuracy \\
\bottomrule
\end{tabular}
\caption{Dataset descriptions and statistics.}\label{tab:benchmark}
\vspace{-0.4cm}
\end{table}

\subsection{Experiments Setup}
\paragraph{Dataset and non-IID partitioning} 
Our experiments use six datasets from the GLUE benchmark. The reasons are as follows: (1) the GLUE datasets have emerged as the \textit{de facto} standard benchmark for assessing PLMs' effectiveness in natural language understanding; (2) They have been extensively leveraged to validate various PETuning methods~\cite{li-liang-2021-prefix,DBLP:journals/corr/abs-2110-07602,zaken2021bitfit,houlsby2019parameter}, and (3) these datasets are large enough and convenient for FL data partitioning (non-IID). Due to the limitation of the unpublished test set in GLUE, we follow the previous studies~\cite{liu2022no,zhao2022reduce} and use the original validation set as the new test set and split a part of the training set as the validation set. The data breakdown for this benchmark is in Table~\ref{tab:benchmark}.

For the non-IID data partitioning, we follow \citet{lin2021fednlp} and partition the datasets by using the Dirichlet distribution as the class priors. In particular, we sample $\mathcal{D} \sim Dir(\alpha)$ and allocate data $\mathcal{D}_k$ to $k$-th client. $\alpha$ determines the degree of non-IID, and a lower value of $\alpha$ generates a high label distribution shift. Unless otherwise specified, we maintain a default setting of $\alpha$ = 1.0 for the Dirichlet parameter throughout our experiments.

\paragraph{Implementation Details} 
We adopt the benchmark FL system FedAvg to simulate the FL setting, which has been applied to commercial products~\cite{bonawitz2019towards}. In FedAvg, clients upload local model parameters to the central server and download the aggregated model for the next training round. Following \citet{lin2021fednlp}, we set the communication round to 100 and the local training epoch to 1 for all tuning methods. Following \citet{chen2022revisiting}, we utilize Roberta-Base~\cite{liu2019roberta} as the local model released by Huggingface\footnote{https://github.com/huggingface/transformers}. The FedAvg implementation is based on \texttt{FedLab}~\cite{JMLR:v24:22-0440}.

\begin{table*}[ht]
    \centering
    \small

    \begin{tabular}{l|cccccc|lrl}
    \toprule
    Methods  & RTE & MRPC & SST-2 & QNLI & QQP & MNLI & Avg. & \multicolumn{1}{c}{Rel.} & Com. \\
    \midrule
    FedFT  & $\bm{70.3_{1.2}}$ & $\bm{90.7_{0.3}}$ & $\bm{94.0_{0.6}}$ & $\bm{91.0_{0.4}}$  & $\bm{89.5_{0.1}}$ & $\bm{86.4_{0.2}}$ & $\bm{83.1}$ & $100.0$\% & $1$x\\
    FedAP & $69.4_{2.6}$ & $89.1_{1.2}$  &  $93.3_{0.6}$ & $90.9_{0.4}$  & $88.4_{0.2}$ & $86.0_{0.4}$ & $82.4$
     & \yblue{$99.1$\%} & \ygreen{$\uparrow$}$60$x \\
    FedLR & $67.4_{4.2}$ & $84.5_{4.5}$  &  $93.6_{0.5}$ & $90.8_{0.3}$  & $87.4_{0.3}$ & $84.9_{0.4}$ & $81.0$
     & \yblue{$97.5$\%} & \ygreen{$\uparrow$}$141$x \\
    FedPF & $58.6_{2.2}$ & $86.8_{1.0}$ &  $93.0_{0.6}$ & $87.6_{0.5}$  & $85.7_{0.3}$ & $82.2_{0.3}$ & $78.4$
     & $94.3$\% & \ygreen{$\uparrow$}$12$x \\
    FedBF & $61.4_{1.7}$ & $84.6_{2.7}$ &  $92.5_{0.7}$ &  $87.2_{0.5}$ & $84.5_{0.5}$ & $81.7_{0.2}$ & $77.8$ 
     & $93.6$\% & \ygreen{$\uparrow$}$190$x \\

    \midrule
    CenFT  & $73.0_{1.4}$ & $90.9_{0.6}$ &  $92.9_{0.2}$ & $90.8_{0.5}$ & $\bm{91.1_{0.2}}$ & $86.0_{0.2}$ & $83.6$ & $100.0$\% & \multicolumn{1}{c}{-} \\
    CenAP & $\bm{76.0_{1.8}}$  & $90.6_{0.8}$ &  $\bm{94.6_{0.5}}$ & $\bm{92.9_{0.1}}$ & $\bm{91.1_{0.1}}$ & $\bm{87.5_{0.2}}$ & $\bm{84.7}$
     & \yred{$101.3$\%} & \multicolumn{1}{c}{-} \\
    CenLR & $74.4_{2.4}$  & $\bm{91.7_{0.6}}$  & $94.0_{0.4}$ & $92.7_{0.6}$ & $90.1_{0.3}$ & $87.0_{0.2}$ & $84.4$ 
     & \yred{$100.9$\%} & \multicolumn{1}{c}{-} \\
    CenPF & $65.6_{5.1}$  & $90.2_{0.9}$  &  $93.7_{0.8}$ & $91.5_{0.2}$ & $89.5_{0.1}$ & $86.7_{0.2}$ & $82.2$ 
     & \yblue{$98.3$\%} & \multicolumn{1}{c}{-} \\
    CenBF  & $70.9_{1.0}$  & $91.3_{0.8}$  &  $94.1_{0.3}$ & $91.3_{0.2}$ & $87.4_{0.2}$ & $84.6_{0.1}$ & $82.6$
     & \yblue{$98.8$\%} & \multicolumn{1}{c}{-} \\
    \bottomrule
    \end{tabular}

    \caption{
    Performance results of the PETuning and FT on GLUE benchmark under the federated (upper half) and the centralized (bottom half) settings. With significantly reducing communication overhead, FedPETuning still maintains acceptable performance.
    The \textit{Rel.} denotes the percentage of PETuning in terms of performance relative to FT. The \yblue{blue} value indicates more than 95\% performance of FT, and \yred{red} value indicates performance in excess of FT. The \textit{Com.} denotes the normalized values of communication overhead of FedPETuning and FedFT. Mean and standard deviation are computed over 5 runs.
    }\label{tab:main_results}
\end{table*}

Under the FedAvg training protocol, we primarily evaluate the full fine-tuning (FedFT) and four representative PETuning methods, covering adapter tuning (FedAP), prefix tuning (FedPF), LoRA (FedLR), and BitFit (FedBF). For FedAP, we follow the architecture in \citet{houlsby2019parameter}, which interposes adapter modules into both the multi-head attention module and the feed-forward network in each Transformer layer. We use prefix tuning~\cite{lester2021power} as the representative of prompt tuning because it has better performance on the small PLMs, e.g., base versions of Roberta. For LoRA and BitFit, we take the architectures from their origin papers~\cite{zaken2021bitfit,DBLP:conf/iclr/HuSWALWWC22}. All PETuing methods are based on \texttt{OpenDelta}\footnote{https://github.com/thunlp/OpenDelta}, which is a plug-and-play framework for parameter-efficient tuning.

To make a fair and reasonable comparison, we run a hyperparameter sweep for each dataset and tuning method. We select the best model according to the metric on the validation set and report test set metric. Especially, the learning rate is selected from \{5e-5, 1e-4, 5e-4, 1e-3, 5e-3, 1e-2\}. We search the reduction factor from \{16, 64\} for FedAP, the prompt length from \{8, 16, 64\} for FedPF, and the scaling factor and rank from \{8, 16\} for FedLR. All experiments are done on a server with 8 Nvidia Tesla V100 GPUs with 32GB RAM each.

\begin{figure}[t]
\centering
\includegraphics[width=0.49\textwidth]{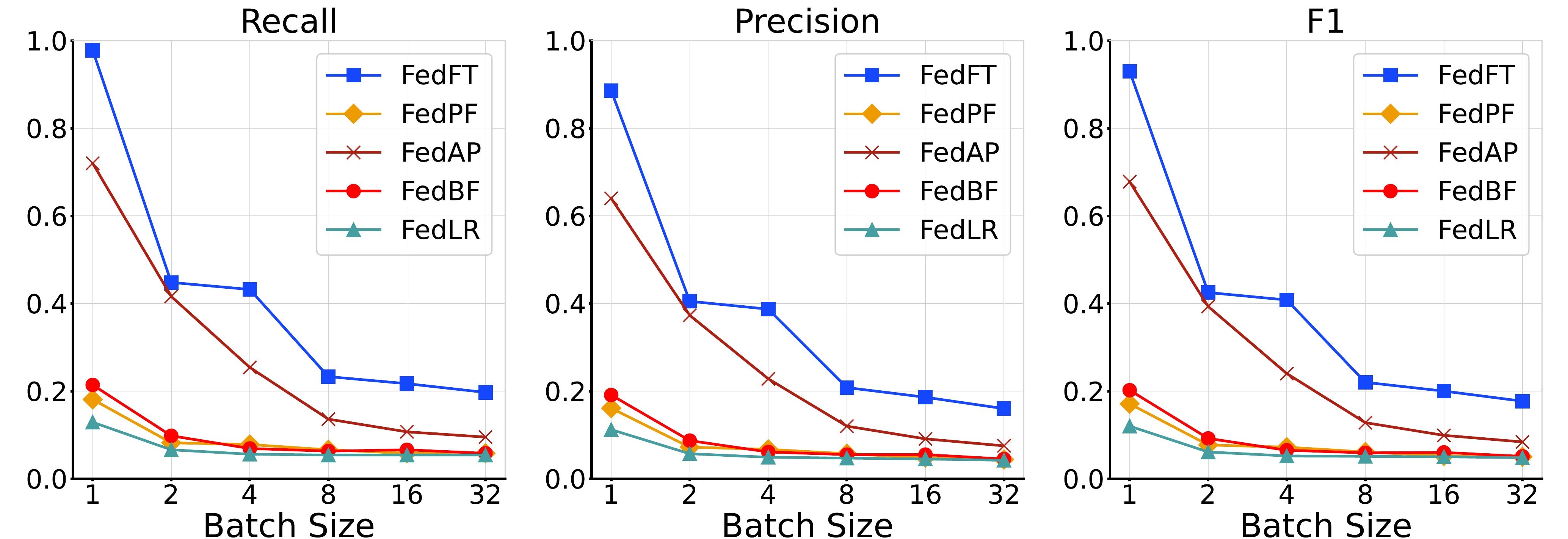}
\caption{The data reconstruction attack results on the attack dataset with different tuning methods. Low performance is better. FedPETuning can effectively defend against data reconstruction attacks.}
\label{fig:dlg}
\end{figure}


\subsection{Privacy-Preserving Results}\label{exp:privacy}
We first investigate the privacy-preserving capabilities of FedPETuing and FedFT. In this privacy attack experiment,  we adopt DLG~\cite{zhu2019deep} as our base attack method. Due to space limitations, we have omitted the working process of DLG and recommend the reader read the original paper~\cite{zhu2019deep}.

\paragraph{Setup and Metrics} As the goal for the attacker with DLG is to recover text from client-uploaded gradients, we follow~\citet{song2020information} and evaluate FedPETuning and FedFT in terms of \textit{precision} (the average percentage of recovered words in the target texts), \textit{recall} (the average percentage of words in the target texts are predicted) and \textit{F1 score} ( the harmonic mean between precision and recall). Specifically, we randomly selected 128 samples from the MNLI as the attack dataset.

\paragraph{Results} Figure~\ref{fig:dlg} shows the results of DLG on the attack dataset with different tuning methods. The results show that \textbf{FedPETuning can effectively defend against data reconstruction attacks compared to FedFT.} Since FedPETuning communicates a fraction of the entire model parameters with the server during the federated training process, it is intractable for the attacker to reconstruct the original text from such lightweight model parameters. Surprisingly, among FedPETuning, DLG is more likely to reconstruct private data from FedAP. We speculate that this result may be related to the design of PETuning methods. Adapter tuning inserts adapter layers to vanilla Transformers, which can encode the input data separately during model training. Compared to other PETuning methods, Adapter tuning is easier to ``remember'' the model input, thus causing more severe privacy leaks.

There is a clear and consistent trend across all tuning methods: DLG performance decreases as the batch size increases. The reason is that the DLG attack requires more parameters to be optimized in larger batch sizes. Therefore, increasing the batch size is a good defense strategy against the DLG attack. However, clients (e.g., mobile phone) in the FL system are usually resource-constrained in real-world applications. There is a trade-off between limited computing resources and large batches. In contrast, the FedPETuning method does not vary significantly across different batch sizes. In this sense, \textit{FedPETuning can provide a practical defense strategy for clients with restricted computing resources.}

\subsection{Performance Comparison}\label{exp:performance}
Table \ref{tab:main_results} shows the performance for four PETuning and FT under the federated (upper half table) and centralized settings (bottom half table). The results demonstrate that \textbf{ FedPETuning maintains acceptable performance (more than 95\% of FedFT's performance) while reducing the communication overhead substantially.}

From the upper half of Table \ref{tab:main_results}, we find that although FedPETuning methods lag behind FedFT in the federated setting, the performance gap is relatively acceptable. For instance,  FedAP reduces 60 times the communication overhead by only 0.7\% performance degradation compared with FedFT. This result also shows that FedPETuning methods, especially FedAP and FedLR (more than 95\% of FedFT's performance), can achieve a good level of trade-off between performance and communication overhead in practice.

In the centralized setting, some PETuning methods achieve competitive performance compared with CenFT. Specifically, CenAP outperforms CenFT on five out of six datasets, and the CenAP and CenLR achieve better performances on average. This result also supports our motivation that 
PETuning, stimulating colossal models with only a small portion of tunable parameters, can naturally be a promising way for FL under communication and resource constraints.

Comparing the upper and bottom parts of Table \ref{tab:main_results}, we find that all tuning methods endure a decline in performance under the federated setting. Remarkably, the FedPETuning exhibits a significant drop in performance in the data heterogeneous FL context, which aligns with the results observed in data heterogeneity experiments (see Section~\ref{exp:niid}). This result suggests that the \textit{FL community may design FL-tailored PETuning methods to bridge the performance gap caused by the non-IID problem.} This will be explored in our future work. 

\begin{figure}[t]
\centering
\includegraphics[width=0.48\textwidth]{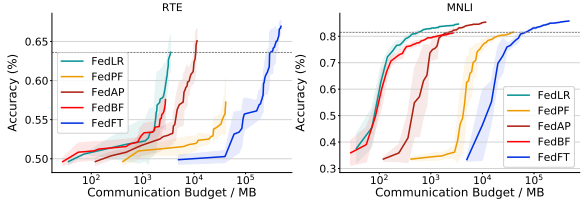}
\caption{Accuracy versus Communication Budget for all tuning methods. The horizontal dashed line indicates the acceptable performance, which is 95\% of the performance of CenFT. FedPETuning makes federated training more efficient by significantly reducing the communication overhead compared to FedFT.}
\label{fig:comm_analysis} 
\end{figure}

\begin{figure}[t]
\centering
\includegraphics[width=0.48\textwidth]{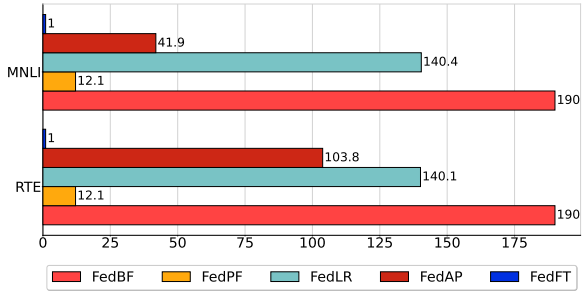}
\caption{Normalized values of storage efficiencies of FedPETuning and FedFT on RTE and MNLI. Higher is better. FedPETuning lowers the storage cost to entry by 12\textasciitilde 190 times.}
\label{fig:para_cost} 
\end{figure}

\subsection{Resource Costs}\label{exp:cost}
We next show the resource cost by different tuning methods under FL settings, including communication budget and client storage overhead. Figure \ref{fig:comm_analysis} shows accuracy versus communication budget for all tuning methods on RTE and MNLI\footnote{The plots of remaining tasks can be found in Appendix \ref{sec:app_extra}, which have similar results.}. As shown in Figure~\ref{fig:comm_analysis}, the communication budgets are ranked as FedFT $\gg$ FedPF $>$ FedAP $>$  FedLR $>$ FedBF. The experimental results show that \textbf{FedPETuning renders federated training remarkably efficient by significantly reducing the communication overhead as opposed to FedFT.} More communication overhead entails increased training time during uploading and downloading. For FL that necessitates high-frequency communication, FedFT is extremely time-consuming, making the federated training slow. In this regard, FedPETuning is more pragmatic in real-world applications, particularly in communication-constrained FL challenges.

Figure \ref{fig:para_cost} shows normalized values of storage efficiencies of FedPETuning and FedFT on RTE and MNLI. \textbf{FedPETuning lowers the storage cost to entry by 12\textasciitilde 190 times.} This appealing characteristic is practiced for local clients of real-world FL systems. When deploying multiple tasks on a local client, FedPETuning can share PLM between different tasks, enabling the client to maintain only a few parameters for each task, thus reducing storage requirements.

\subsection{Impact of Data Heterogeneity}\label{exp:niid} 
As data heterogeneity is a fundamental challenge in FL, we also evaluate the performance of FedPETuning and FedFT under different data heterogeneity. Following \citet{lin2021fednlp}, we consider three Dirichlet distributions in this experiment by choosing $\alpha$ from $\{0.1, 1.0, 10.0\}$ where a smaller $\alpha$ indicates a sharper non-IID distribution among clients. 

\begin{figure}[t]
\centering
\includegraphics[width=0.48\textwidth]{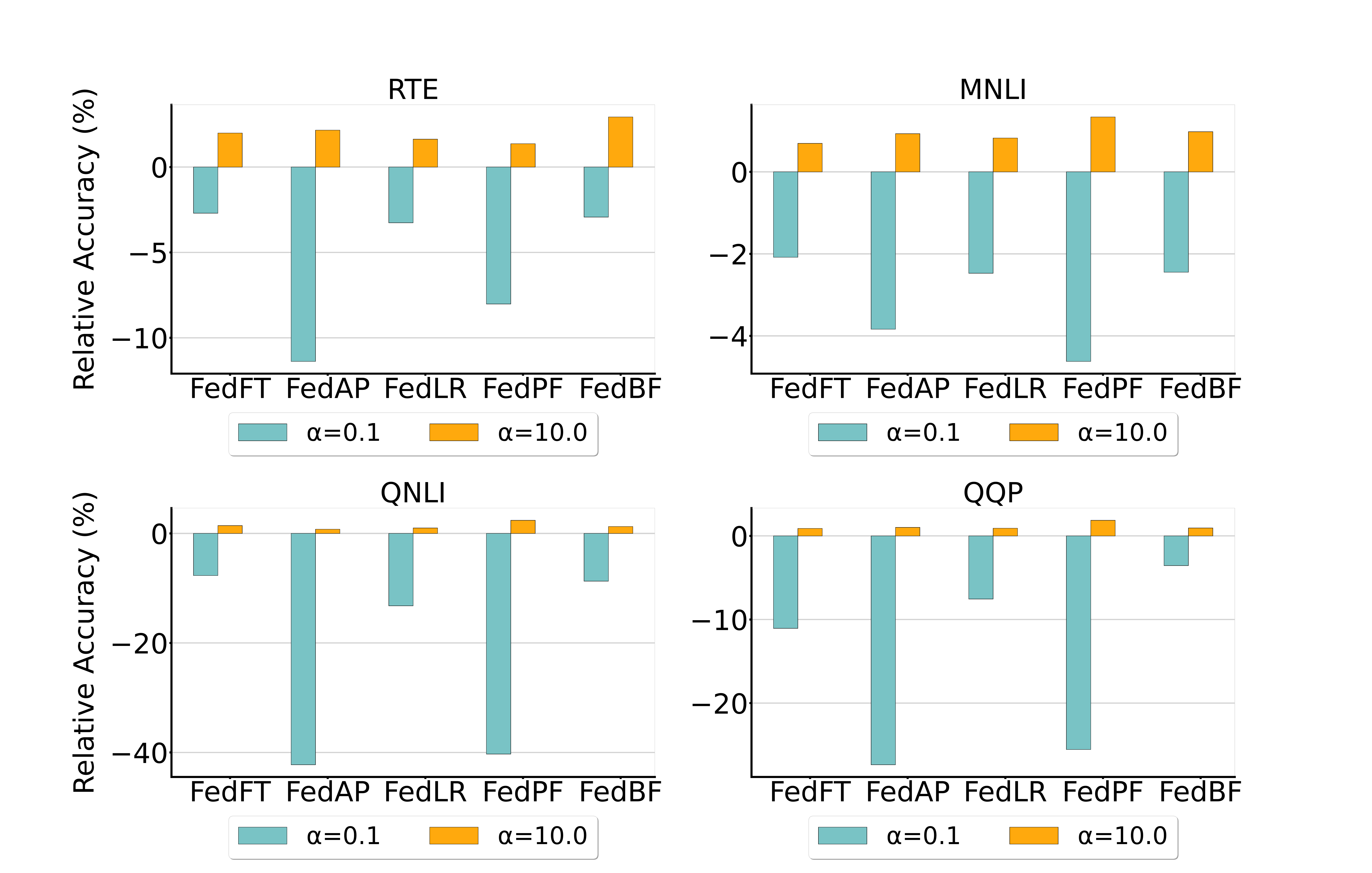}
\caption{Performance variation of FedPETuning and FedFT under different non-IID with respect to $\alpha=1.0$ on RTE and MNLI. FedPETuning is more susceptible to data heterogeneity than FedFT.} 
\label{fig:alpha} 
\end{figure}

\begin{figure}[t]
\centering
\includegraphics[width=0.48\textwidth]{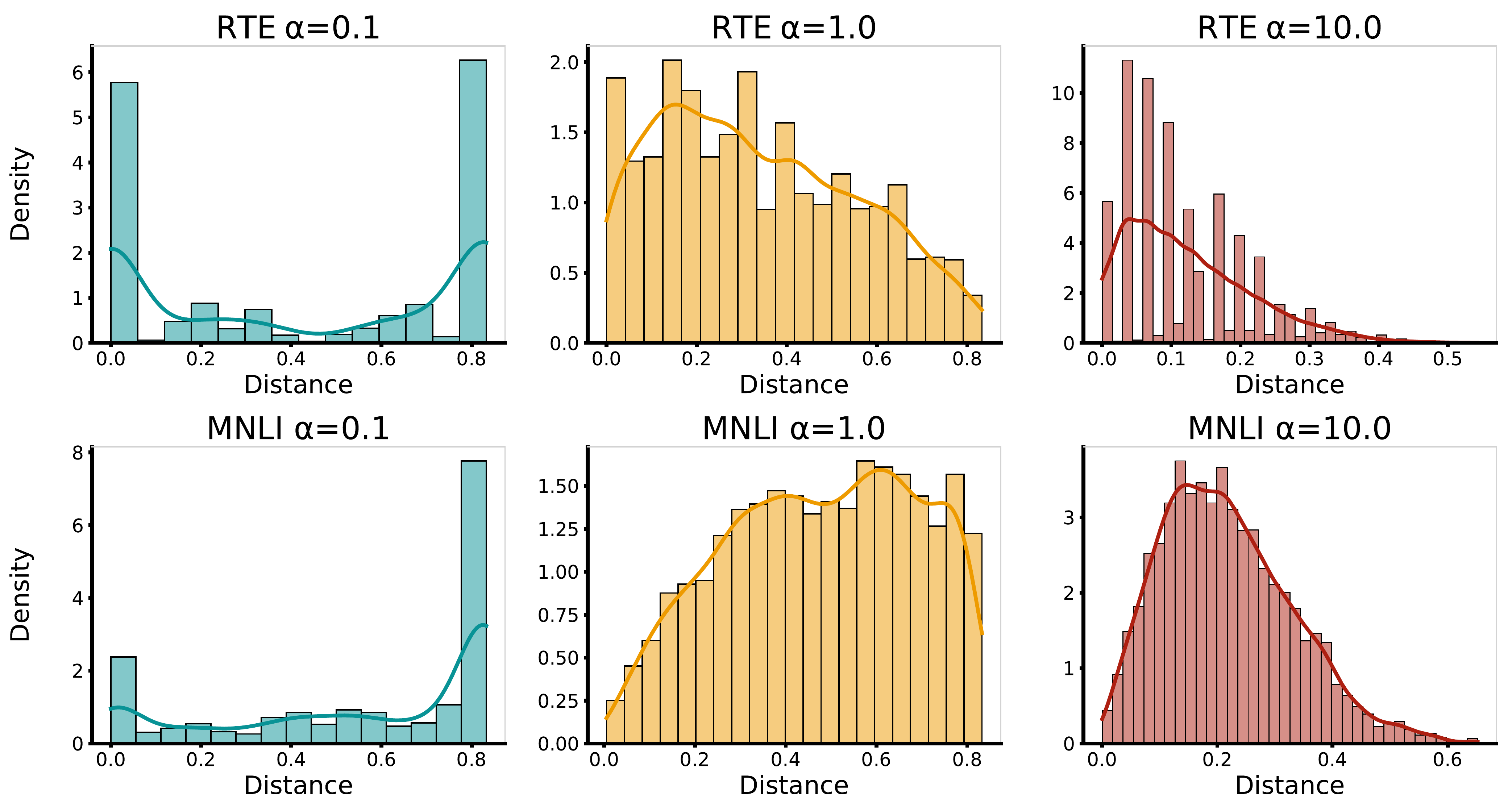}
\caption{Data distribution under different Dirichlet parameter $\alpha$. We report the pairwise Jensen–Shannon distance of the label distribution between two clients.}
\label{fig:distribution} 
\end{figure}

The performance of $\alpha=1.0$ has already been discussed, Figure~\ref{fig:alpha} illustrates performance changes of FedPETuning and FedFT with respect to the other two $\alpha$ values. The outcomes corresponding to other tasks can be found in Appendix \ref{sec:app_extra}. Figure \ref{fig:alpha} reveals that \textbf{FedPETuning is more susceptible to data heterogeneity than FedFT.} As shown in Figure~\ref{fig:alpha}, although the increase in data heterogeneity ($\alpha$ from 1.0 to 0.1) degrades the performance of all tuning methods, FedPETuning degrades more dramatically compared to FedFT. This suggests more the federated model necessitates more trainable parameters to tackle intricate data heterogeneity, and \textit{it may be arduous for FedPETuning to confront complex data heterogeneity in FL.} 

Another noteworthy phenomenon is the performance of FedPETuning and FedFT do not change much when increasing $\alpha$ from 1.0 to 10.0. To figure out this, we report the Probability Density Function (PDF) of the data distribution under different $\alpha$ in Figure~\ref{fig:distribution} (See Appendix \ref{sec:app_partitions} for other datasets). As shown in Figure~\ref{fig:distribution}, the data heterogeneity is similar for $\alpha$ equal to 1.0 and 10.0, while the distribution gap is large for $\alpha$ equal to 0.1. We think similar data heterogeneity between $\alpha=1.0$ and $\alpha=10.0$ contributes to this result. 

\begin{figure}[t]
\centering
\includegraphics[width=0.48\textwidth]{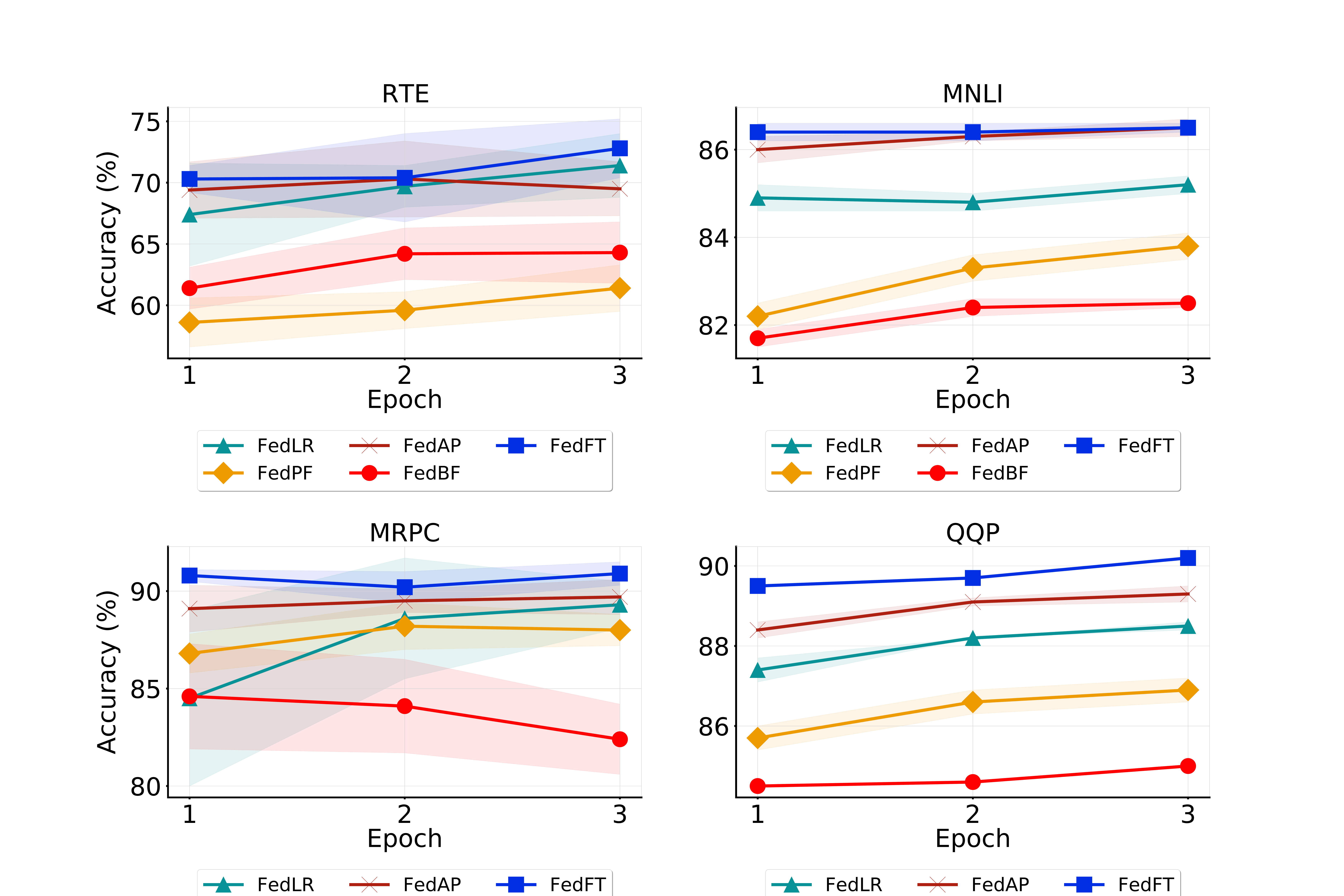}
\caption{Performance comparison of FedPETuning and FedFT under different epochs on RTE and MNLI. Most tuning methods under the FL setting benefit from increased local training epochs.}
\label{fig:epoch} 
\vspace{-0.2cm}
\end{figure}

\subsection{Impact of Local Training Epochs}\label{exp:epoch}
Figure~\ref{fig:epoch} shows the performance of FedPETuning and FedFT with different local training epochs on RTE and MNLI. More results are provided in Appendix~\ref{sec:app_extra}. This result reveals that \textbf{most tuning methods under the FL setting benefit from increased local training epochs.} Federated models trained with more local epochs consistently yield a performance gain across different tuning methods when clients with heigh-resource data (like MNLI). In contrast, training with more epochs may incur a drop in performance for the low-resource dataset (like FedLR and FedAP on RTE). This may be due to the probability of PLMs becoming more susceptible to overfitting when clients with few data undergo more training epochs. On the contrary, clients with adequate data require more local training epochs to enhance the FL performance.

With the data scale into analysis, we also find that \textit{all tuning methods are more unstable on small datasets} in Figure~\ref{fig:epoch}. For instance, the standard deviation performance of FedLR on RTE and MRPC is over $4.0$\% while other datasets are no more than $0.5$\%. This phenomenon is consistent with previous work~\cite{chen2022revisiting}, and they experimentally show that increasing training steps can effectively mitigate the instability of the PETuning. However, this strategy fails in the FL setting. Training PLMs more stably in FL is a good research direction we will explore in our future work.

\begin{table}[t]
    \centering
    \resizebox{\linewidth}{!}{
    \begin{tabular}{cccc|lrl}
    \toprule
    Methods & RTE & MRPC & SST-2 & Avg. & \multicolumn{1}{c}{Rel.} & Com. \\
    \midrule
    FedFT  & $70.4$ & $\bm{90.2}$ & $94.3$ & $\bm{85.0}$ & $100.0$\% & $1$x\\
    FedAP & $\bm{71.5}$ & $88.5$ & $94.0$ & $84.7$ &
    \yblue{$99.7\%$} & \ygreen{$\uparrow$}$70$x \\
    FedLR & $70.8$ & $89.8$ & $\bm{94.4}$ & $\bm{85.0}$ &
    \yred{$100.0\%$} & \ygreen{$\uparrow$}$141$x \\
    FedPF & $66.4$ & $88.1$ & $93.7$ & $82.7$ &
    \yblue{$97.3\%$} & \ygreen{$\uparrow$}$12$x \\
    FedBF &  $55.2$ & $88.6$ & $92.8$ & $78.9$ & 
    $92.8\%$ & \ygreen{$\uparrow$}$189$x \\
    \bottomrule
    \end{tabular}
    }
    \caption{Performance results of the FedPETuning and FedFT in the cross-silo setting. Significantly reducing the communication overhead, FedPETuning still achieves acceptable performance in the  cross-silo FL scenario.} \label{tab:cross-silo}
    \vspace{-0.2cm}
\end{table}

\begin{table}[t]
    \centering
    
    \resizebox{\linewidth}{!}{
    \begin{tabular}{cccc|lrl}
    \toprule
    Methods & QNLI & QQP & MNLI & Avg. & \multicolumn{1}{c}{Rel.} & Com. \\
    \midrule
    FedFT & $\bm{87.9}$ & $\bm{88.8}$ & $\bm{85.6}$ & $\bm{87.4}$ & $100.0$\% & $1$x\\
    FedAP & $85.9$ & $87.0$ & $84.9$ & $85.9$ &
    \yblue{$98.3\%$} & \ygreen{$\uparrow$}$52$x \\
    FedLR & $86.0$ &  $86.5$ & $84.7$ & $85.7$ &
    \yblue{$98.1\%$} & \ygreen{$\uparrow$}$140$x \\
    FedPF & $84.6$ & $81.8$ & $80.4$ & $82.3$ &
    $94.2\%$ & \ygreen{$\uparrow$}$12$x \\
    FedBF & $80.5$ & $84.0$ & $80.7$ & $81.7$ &
    $93.5\%$ & \ygreen{$\uparrow$}$190$x \\
    \bottomrule
    \end{tabular}
    }
    \caption{Performance results of FedPETuing and FedFT in the large-scale cross-device setting. With significantly reducing the communication overhead, FedPETuning still achieves acceptable performance in large-scale cross-device FL scenarios.}  \label{tab:large-scale}
    \vspace{-0.2cm}
\end{table}

\subsection{Various FL Scenarios}\label{exp:scenario}
To verify the effectiveness of PETuning under different FL scenarios, we mainly consider two FL scenarios in this experiment, i.e., the cross-silo FL scenario~\cite{kairouz2021advances} and the large-scale cross-device FL scenario~\cite{lai2022fedscale}. Cross-silo~\cite{kairouz2021advances} is a vital application scenario for FL, suitable for several users (no more than 100). In Cross-silo, the server selects all clients for training in each communication round. Large-scale cross-device FL ~\cite{lai2022fedscale} is another federated scenario for deployment across thousands of clients. In the large-scale cross-device FL, data held by the local client is more scarce. 
In this experiment, we chose RTE, MRPC, and SST-2 to simulate the cross-silo FL scenarios, while QNLI, QQP, and MNLI to simulate the large-scale cross-device FL scenarios. The total number of clients in the cross-silo and large-scale settings is set to 10 and 1000, respectively. For both FL scenarios, ten clients are involved in each communication round for local training and federated aggregation. 

Table~\ref{tab:cross-silo} and Table~\ref{tab:large-scale} show the performance results of PETuning and FT under these two FL scenarios, respectively. The results show that \textbf{FedPETuning can  significantly reduce the communication overhead while achieving acceptable performance in cross-silo and large-scale cross-device FL scenarios.} More noteworthy between the different FL scenes is the cross-silo setting. As shown in Table~\ref{tab:cross-silo}, the performance gap between the FedPETuning and FedFT diminishes under the cross-silo setting, compared to our standard setting in Section~\ref{exp:performance}. For example, FedPF lags behind FedFT by $2.3\%$ and $4.7\%$ under the cross-silo FL scenario and the FL scenario of Table~\ref{tab:main_results}, respectively. We attribute this observation to relieving the data scarcity issue since each client has ten times more training samples in the cross-silo setting than in the standard setting.

\section{Conclusion}
\label{sec:conclu}
This paper investigates parameter-efficient tuning methods of PLMs in the FL setting with extensive experiments for in-depth measurement of these methods under FL settings, covering privacy attacks, performance comparisons, and resource-constrained analysis. Experimental results unveil that FedPETuning can (1) achieve acceptable performance while reducing the colossal communication overhead and local storage cost, and (2) provide strong privacy-preserving capacity under different FL settings. To facilitate FL-tailored PETuning research, we have released our code and partitioned datasets, aiming to facilitate the use of FedPETuning and inspire the broader community to develop more suitable PETuning methods in future federated learning research.

\section*{Limitation}
\label{sec:limited}
One limitation of this paper is that we do not validate FedPETuning on large scale models, (e.g., T5~\cite{DBLP:journals/corr/abs-1910-10683}, LLaMa~\cite{touvron2023llama}, Vicuna~\cite{vicuna2023}, etc). Although the parameter efficiency of PETuning is more impressive on larger pre-trained models, we have to consider the limited computational resources of clients in FL, making it challenging to deploy such a large-scale model.  In addition, with the increasing size of modern pre-trained models, the community needs to design FL-friendly PETuning methods. In this sense, our work can serve as a benchmark and guide for future exploration of PETuning in FL.

\section*{Acknowledgements}
We'd like to thank all the anonymous reviewers for their careful readings and valuable comments. This work was partially supported by the National Key Research and Development Program of China (No. 2018AAA0100204), a key program of fundamental research from Shenzhen Science and Technology Innovation Commission (No. JCYJ20200109113403826), the Major Key Project of PCL (No. 2022ZD0115301), and an Open Research Project of Zhejiang Lab (NO.2022RC0AB04).

\bibliography{anthology,EX_Custom}
\bibliographystyle{acl_natbib}

\clearpage
\appendix


\section{Non-IID Partitionings Results}\label{sec:app_partitions}
\begin{figure}[t]
\centering
\includegraphics[width=0.48\textwidth]{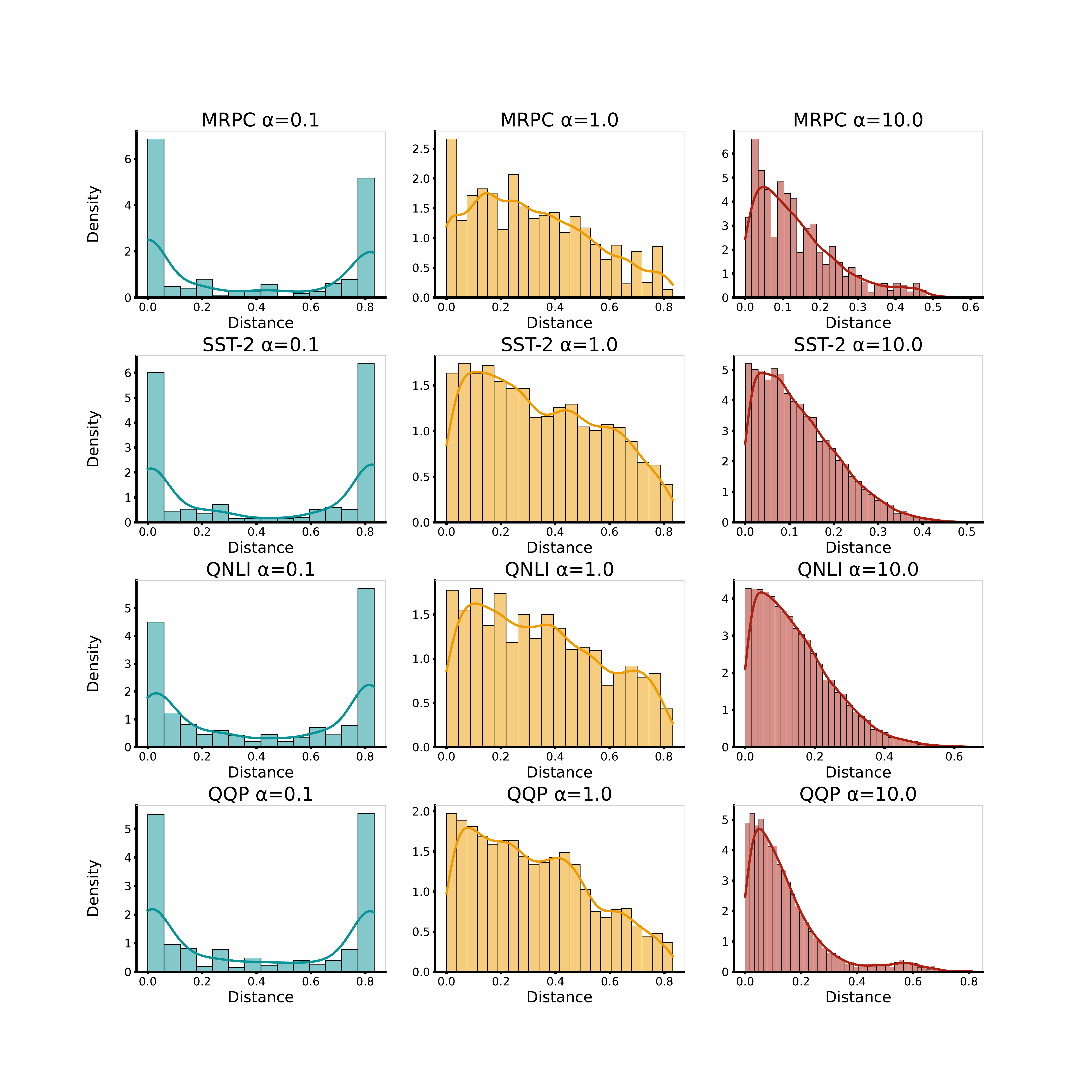}
\caption{Data distribution under different $\alpha$. We report the pairwise Jensen–Shannon distance of the label distribution between two clients.}
\label{fig:appen_distribution} 
\vspace{-0.4cm}
\end{figure}

Figure~\ref{fig:appen_distribution} illustrates the data distribution under different $\alpha$. 
It is observed that data distributed in  clients with $\alpha$ as $0.1$ has a large distance with each other, which has impact on the performance.
Both $\alpha$ as 1.0 and 10.0 are considered to produce a more uniform distribution.


\section{Extra Results}\label{sec:app_extra}
\begin{figure}[]
\centering
\includegraphics[width=0.48\textwidth]{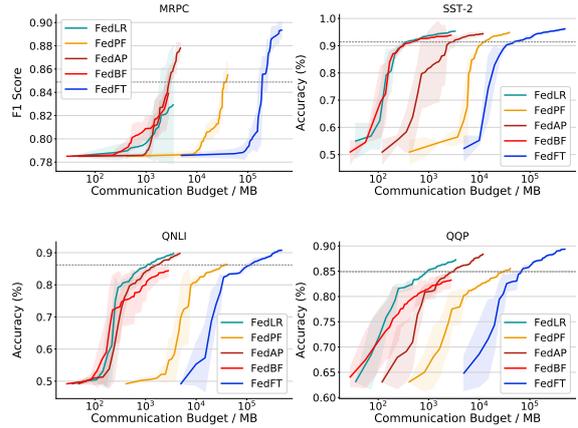}
\caption{Accuracy versus Communication Budget for all tuning methods. The horizontal dashed line indicates the acceptable performance, which is 95\% of the accuracy of CenFT.}
\label{fig:appen_comm_analysis}
\vspace{-0.4cm}
\end{figure}

\paragraph{Communication Analysis}
In this section, we illustrate the accuracy given the communication budget on the remaining four tasks in Figure \ref{fig:appen_comm_analysis}.
As can be seen, PETuning methods consistently reduce the communication budget over several orders of magnitude while providing comparable performance.
Moreover, PETuning methods (apart from FedBF) achieve acceptable accuracy ( 90\% of fine-tuning) on all the tasks, demonstrating the effectiveness of these methods.

\begin{figure}[t]
\centering
\includegraphics[width=0.45\textwidth]{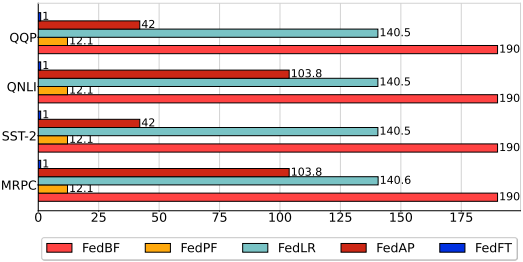}
\caption{Normalized values of storage efficiencies of
FedPETuning and FedFT on the other four tasks. Higher is better.}
\label{fig:appen_para_cost} 
\vspace{-0.4cm}
\end{figure}

\begin{figure}[]
\centering
\includegraphics[width=0.48\textwidth]{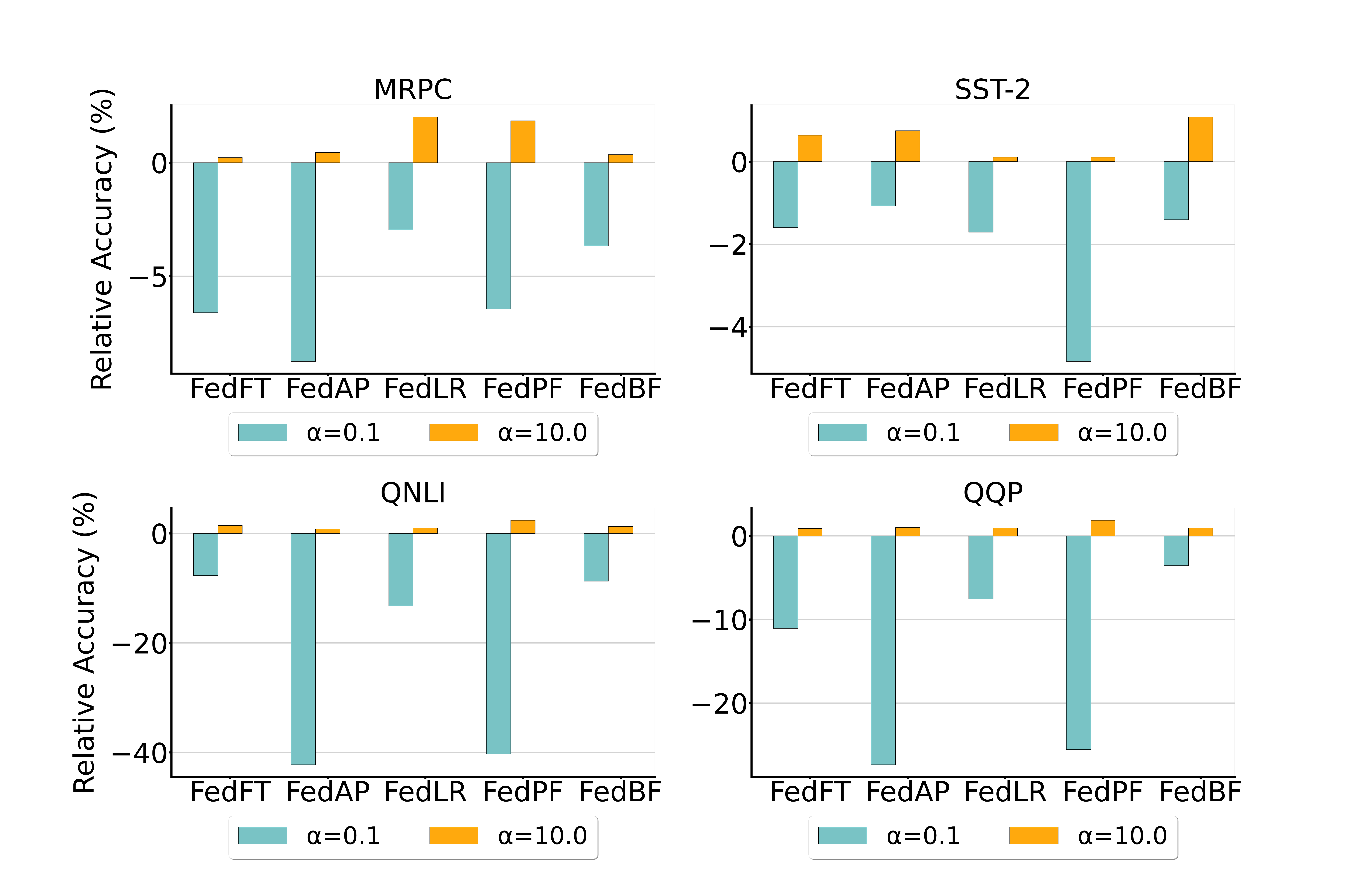}
\caption{Relative performance of FedPETuning and FedFT on RTE and MNLI under different Dirichlet distributions.} 
\label{fig:appen_alpha_analysis}
\vspace{-0.3cm}
\end{figure}

\begin{figure}[]
\centering
\includegraphics[width=0.48\textwidth]{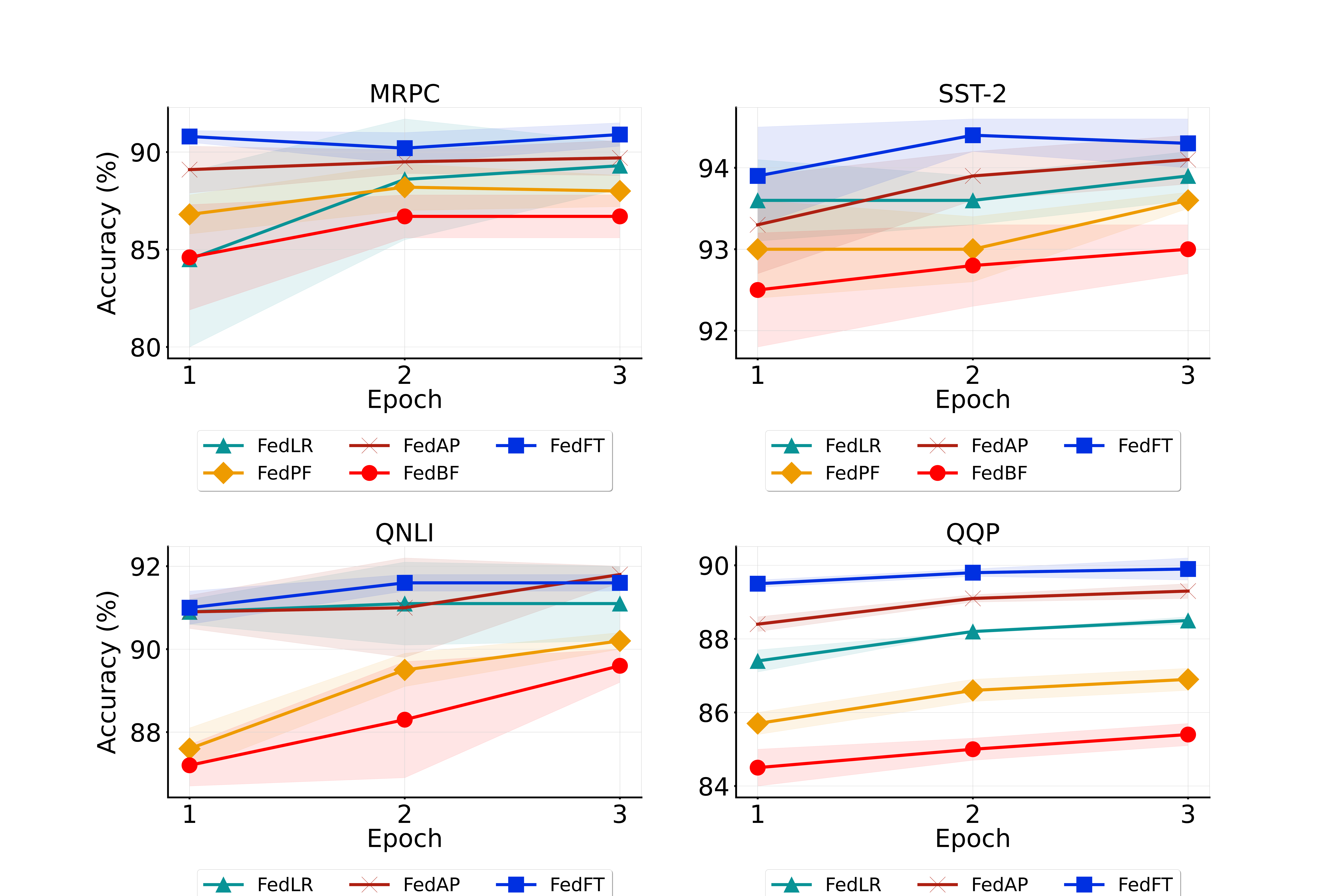}
\caption{Accuracy comparison of PETuning methods and FT under different epochs.}
\label{fig:appen_epoch_analysis}
\vspace{-0.4cm}
\end{figure}

\paragraph{Non-IID Analysis} Figure \ref{fig:appen_alpha_analysis} presents the comparisons of accuracy under different distributions by varying $\alpha$.
All the methods obtain better performance with $\alpha$ as $1.0$ than $\alpha$ as $0.1$, showing that non-IID distribution has a negative impact on the model's performance.
On the contrary, varying $\alpha$ from $1.0$ to $10.0$ has little impact in most circumstances.

\paragraph{Local Training Epoch Analysis} 
In Figure \ref{fig:appen_epoch_analysis}, we show the accuracy of tuning methods trained with different numbers of training epochs on the local clients. The accuracy of the relatively small datasets (MRPC, SST-2) shows a faint decreasing trend because of the over-fitting issue.
All the tuning methods benefit from more training epochs on relatively large datasets (QNLI, QQP).

\end{document}